\documentclass[conference]{IEEEtran}
\pdfoutput=1

\IEEEoverridecommandlockouts
\usepackage{cite}
\usepackage{amsmath,amssymb,amsfonts}
\usepackage{algorithmic}
\usepackage{graphicx}
\usepackage{textcomp}
\usepackage{xcolor}
\usepackage{booktabs}
\usepackage{hyperref}
\usepackage{subcaption}
\usepackage{multirow}
\def\BibTeX{{\rm B\kern-.05em{\sc i\kern-.025em b}\kern-.08em
    T\kern-.1667em\lower.7ex\hbox{E}\kern-.125emX}}
\begin{document}

\title{Deep Learning for Taxol Exposure Analysis: A New Cell Image Dataset and Attention-Based Baseline Model}

\author{\IEEEauthorblockN{1\textsuperscript{st} Sean Fletcher}
\IEEEauthorblockA{\textit{Computer Science Department} \\
\textit{University of Southern Maine}\\
Portland, United States \\
sean.fletcher@maine.edu}
\and
\IEEEauthorblockN{2\textsuperscript{nd} Gabby Scott}
\IEEEauthorblockA{\textit{Biology Department } \\
\textit{University of Southern Maine}\\
Portland, United States \\
gabrielle.scott1@maine.edu}
\and
\IEEEauthorblockN{3\textsuperscript{rd} Douglas Currie}
\IEEEauthorblockA{\textit{Biology Department } \\
\textit{University of Southern Maine}\\
Portland, United States \\
douglas.currie@maine.edu}
\and 
\IEEEauthorblockN{4\textsuperscript{th} Xin Zhang}
\IEEEauthorblockA{\textit{Computer Science Department} \\
\textit{University of Southern Maine}\\
Portland, United States \\
xin.zhang@maine.edu}
\and
\IEEEauthorblockN{5\textsuperscript{th} Yuqi Song}
\IEEEauthorblockA{\textit{Computer Science Department} \\
\textit{University of Southern Maine}\\
Portland, United States \\
yuqi.song@maine.edu}
\and
\IEEEauthorblockN{6\textsuperscript{th} Bruce MacLeod}
\IEEEauthorblockA{\textit{Computer Science Department} \\
\textit{University of Southern Maine}\\
Portland, United States \\
macleod@maine.edu}
}

\maketitle

\begin{abstract}
Monitoring the effects of the chemotherapeutic agent Taxol at the cellular level is critical for both clinical evaluation and biomedical research. 
However, existing detection methods require specialized equipment, skilled personnel, and extensive sample preparation, making them expensive, labor-intensive, and unsuitable for high-throughput or real-time analysis. 
Deep learning approaches have shown great promise in medical and biological image analysis, enabling automated, high-throughput assessment of cellular morphology.
Yet, no publicly available dataset currently exists for automated morphological analysis of cellular responses to Taxol exposure.
To address this gap, we introduce a new microscopy image dataset capturing C6 glioma cells treated with varying concentrations of Taxol. 
To provide an effective solution for Taxol concentration classification and establish a benchmark for future studies on this dataset, we propose a baseline model named ResAttention-KNN, which combines a ResNet-50 with Convolutional Block Attention Modules and uses a k-Nearest Neighbors classifier in the learned embedding space.
This model integrates attention-based refinement and non-parametric classification to enhance robustness and interpretability. 
Both the dataset and implementation are publicly released to support reproducibility and facilitate future research in vision-based biomedical analysis.
\end{abstract}


\section{Introduction}
Paclitaxel, commercially known as Taxol, is a widely used chemotherapeutic agent effective against a variety of cancers~\cite{klein2021pathomechanisms}. While its clinical significance is well-established, detecting and monitoring Taxol concentrations is critical from two important perspectives. First, for human and animal health, exposure to excessive Taxol—whether through improper handling, contamination, or environmental leakage—can lead to severe toxicological effects such as neurotoxicity and immunosuppression. Second, from an environmental standpoint, Taxol residues released into water systems or soil can disrupt biological processes and pose ecological risks to non-target organisms~\cite{li2021anticancer, d2023toxic}.
Therefore, accurate and timely identification of Taxol presence and concentration is crucial for both biosafety and ecological preservation.

Traditionally, Taxol detection relies on manual laboratory-based techniques such as high-performance liquid chromatography (HPLC) or mass spectrometry~\cite{posocco2018new}. While these methods provide high precision, they are expensive, time-consuming, and require skilled personnel, limiting their accessibility and scalability~\cite{briki2024liquid}. Moreover, visual inspection of Taxol-related samples—such as those derived from extraction or synthesis processes—is subjective and error-prone, especially when minor differences in appearance signal significant chemical variation. These challenges highlight the need for an automated, accessible, and efficient solution~\cite{see2012visual}.

To address these limitations, we explore vision-aided Taxol concentration classification using computer vision models. However, a major barrier to advancing this approach is the lack of publicly available datasets containing labeled images of Taxol samples at varying concentration levels. To address this gap, we construct and release a new dataset composed of Taxol-related images across multiple concentration levels. The dataset comprises 438 high-resolution images categorized into 4 classes based on known concentrations (e.g., Control, $20\mu M$, $40\mu M$, $100\mu M$). Each image is labeled and quality-checked to ensure its utility for supervised learning tasks.

In addition, we provide a baseline model to facilitate future work in this area. Our proposed model, ResAttention-KNN, combines the representational power of ResNet-50~\cite{he2016deep} with the Convolutional Block Attention Module (CBAM)~\cite{woo2018cbam} to enhance spatial and channel-wise feature learning. We further integrate K-Nearest Neighbors (k-NN)~\cite{guo2003knn} as a post-processing classifier to leverage feature similarity for final prediction, which helps improve accuracy in this fine-grained classification task~\cite{srinivas2019hybrid}.

In summary, our contributions are three-fold:
\begin{itemize}
    \item We construct and release a labeled dataset containing high-resolution images of Taxol samples at different concentration levels. This is \textbf{the first dataset} of its kind and provides a foundation for vision-based Taxol concentration analysis.
    \item We propose a baseline model that combines ResNet-50, CBAM attention, and k-NN for effective feature extraction and classification. This model achieves strong performance and can serve as a benchmark for future research.
    \item We make both the dataset\footnote{\url{https://github.com/taxol-image-classification/dataset}} and model implementation\footnote{\url{https://github.com/SeanEdwardFletcher/Microscopy_Image_Classification}} \textbf{publicly available} to support reproducibility and community engagement. This enables other researchers to build upon our work and explore new methods in this domain.
\end{itemize}

The rest of the paper is organized as follows. Section~\ref{sec:rw} discusses the related work. Section~\ref{sec:ds} provides a detailed description of the proposed Taxol dataset. Our proposed method is presented in Section~\ref{sec:pm}. Section~\ref{sec:exp} shows the experimental settings and results.  Finally, conclusions are drawn in Section~\ref{sec:con}.

\section{Related Work}
\label{sec:rw}
To the best of our knowledge, there is currently no existing work that uses computer vision to classify or detect Taxol concentrations based on visual characteristics. From our understanding, we are the first to explore a vision-aided approach for Taxol concentration classification using supervised image-based modeling.

In contrast, the use of computer vision for classification and diagnostic tasks in medical and biological domains is well established. Deep learning models have been developed for detecting pneumonia from chest X-ray images, achieving performance comparable to trained radiologists~\cite{rajpurkar2018deep}. Convolutional neural networks have also been applied to classify skin lesions from clinical images, reaching dermatologist-level accuracy~\cite{esteva2017dermatologist}. In histopathology, deep neural networks have been used for mitosis detection in breast cancer tissue, significantly improving diagnostic automation~\cite{cirecsan2013mitosis}. Similarly, attention-based CNNs have been employed to classify arsenic exposure levels in neuron-like cell images, demonstrating the effectiveness of vision models in capturing subtle biological variations~\cite{maarefdoust2024attention}.


\section{Dataset}
\label{sec:ds}
Due to the lack of publicly available datasets for vision-based Taxol concentration analysis, we constructed a new image dataset to support this task. It features C6 glioma cells—a rat-derived cell line widely used in brain cancer research—exposed to varying concentrations of Taxol~\cite{chao2015induction}. These cells, originating from chemically induced glial tumors in Wistar rats, are commonly used as an in vitro model for glioblastoma, owing to their high proliferation rate and expression of glial fibrillary acidic protein (GFAP)~\cite{giakoumettis2018c6}. Their biological relevance makes them well-suited for studying cellular responses to chemotherapeutic agents in controlled settings. The dataset is publicly available at: \url{https://github.com/taxol-image-classification/dataset}.


\subsection{Image Acquisition}

C6 glioma cells were cultured in Kaighn’s modification of Ham’s F-12 medium (F12K) supplemented with 2.5\% fetal bovine serum (FBS), 15\% horse serum, and 1\% Penicillin-Streptomycin (PenStrep), and maintained at $37^\circ$C in a humidified atmosphere of 5\% CO\textsubscript{2} and 95\% air. Cells were seeded into 24-well plates and treated with Taxol dissolved in dimethyl sulfoxide (DMSO) at final concentrations of 100~\textmu M, 40~\textmu M, and 20~\textmu M. A DMSO-only control group was also included. Cells were incubated under these conditions for 72 hours. Separate plates were fixed using paraformaldehyde. Phase contrast microscopy was performed at 20$\times$ magnification to capture $\sim\!8$ images from each well.

\begin{figure}[htbp]
    \centering
    \includegraphics[width=\columnwidth]{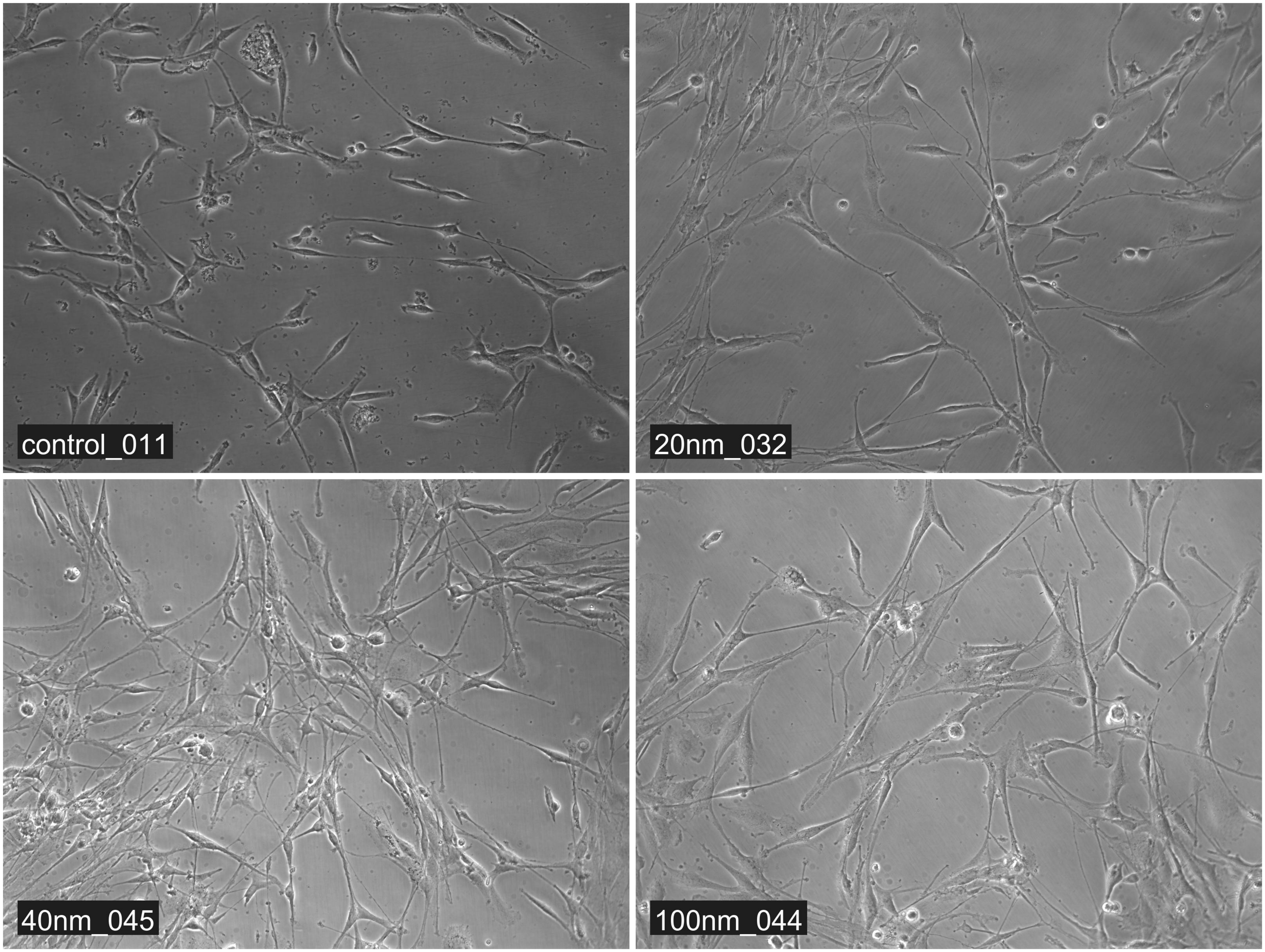}
    \caption{Representative images from each class: Control, 20~\textmu{}M, 40~\textmu{}M, and 100~\textmu{}M Taxol. Each panel displays a single example drawn from the corresponding treatment group.}
    \label{fig:combined_image}
\end{figure}

The experimental conditions included the following four groups:
\begin{itemize}
    \item \textbf{Control} (DMSO only): 109 images
    \item \textbf{20 \textmu{}M Taxol} in DMSO: 109 images
    \item \textbf{40 \textmu{}M Taxol} in DMSO: 110 images
    \item \textbf{100 \textmu{}M Taxol} in DMSO: 110 images
\end{itemize}
Representative examples from each treatment group are shown in Fig.~\ref{fig:combined_image}, including Control, 20~\textmu{}M, 40~\textmu{}M, and 100~\textmu{}M Taxol-treated cells.

All images were captured as grayscale at a resolution of 1600~$\times$~1200 pixels and saved in \texttt{.jpg} format.

\subsection{Data Splits}

To train and evaluate deep learning models, the dataset was split into training, validation, and test sets. Each class contributed 16 images to both the validation and test sets, with the remaining 77 or 78 images were used for training, Table~\ref{tab:dataset_splits}. The images were randomly shuffled before splitting, and a stratified splitting strategy was used to preserve class proportions. To prevent data leakage, image names and indices were tracked throughout the process to ensure that no image appeared in more than one subset.

\begin{table}[htbp]
\caption{Image distribution across splits by class.}
\centering
\begin{tabular}{lccc}
\toprule
\textbf{Class} & \textbf{Train} & \textbf{Validation} & \textbf{Test} \\
\midrule
Control (DMSO only)     & 77 & 16 & 16 \\
20~\textmu{}M Taxol             & 77 & 16 & 16 \\
40~\textmu{}M Taxol             & 78 & 16 & 16 \\
100~\textmu{}M Taxol            & 78 & 16 & 16 \\
\midrule
\textbf{Total}          & 310 & 64 & 64 \\
\bottomrule
\end{tabular}
\label{tab:dataset_splits}
\end{table}


\section{Proposed Model: ResAttention-KNN}
\label{sec:pm}
\begin{figure*}[htbp]
    \centering
    \includegraphics[width=.95\linewidth]{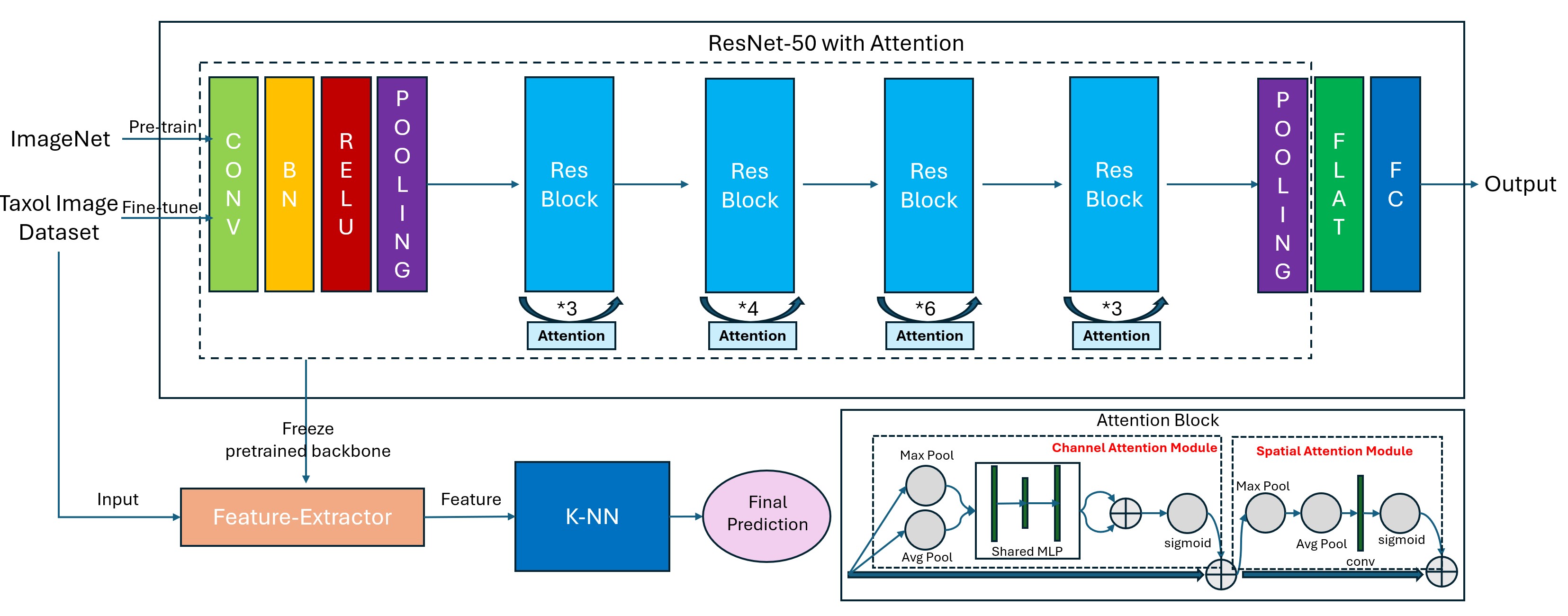} 
    \caption{Pipeline of the proposed ResAttention-KNN model. A ResNet-50 backbone pretrained on ImageNet is enhanced with Convolutional Block Attention Modules (CBAM) after each residual stage to refine feature representations. The backbone is frozen during training and used as a fixed feature extractor. The extracted features are passed to a K-Nearest Neighbors (K-NN) classifier to predict Taxol concentration levels.}
    \label{fig:pp}
\end{figure*}

To effectively classify our newly constructed Taxol concentration dataset, we propose a model named ResAttention-KNN, which will serve as a baseline for future research. For more details on the implementation and hyperparameter selection, please refer to \url{https://github.com/SeanEdwardFletcher/Microscopy_Image_Classification}. 

Our proposed model, ResAttention-KNN, combines a pretrained ResNet-50 with attention modules to better capture important visual patterns in the images. We fine-tune the backbone on the Taxol concentration dataset to adapt it to our specific task, then use it to extract meaningful features. These features are passed to a K-Nearest Neighbors (K-NN) classifier for the final prediction. This approach is both simple and effective for classifying Taxol concentration levels. The pipeline of our approach is shown in Fig.~\ref{fig:pp}.

We choose ResNet-50~\cite{he2016deep} as the base architecture for our proposed approach due to its strong feature extraction capabilities in image classification tasks. Its residual connections enable efficient training of deep networks and help capture complex visual patterns~\cite{yosinski2014transferable}. Furthermore, using a version pretrained on ImageNet allows us to leverage transferable visual knowledge, which is especially beneficial when working with limited training data~\cite{krizhevsky2012imagenet}.

\subsection{Attention Mechanism}
To further enhance the model’s ability to focus on relevant morphological features, we incorporate the Convolutional Block Attention Module (CBAM)~\cite{woo2018cbam} into the ResNet-50 architecture. CBAM introduces a lightweight and effective attention mechanism that sequentially applies channel attention and spatial attention, allowing the model to refine feature representations by emphasizing informative components and suppressing less relevant ones. This mechanism helps the network better capture subtle yet critical visual patterns, thereby improving both classification performance and interpretability.

In our implementation, CBAM is inserted after the final convolutional layer of each residual block and before the residual addition. This placement ensures that the refined features produced by the attention modules are aligned with the residual learning process, preserving the original ResNet structure while enhancing its representational power. By guiding the model to attend to the most informative spatial and channel-wise cues, CBAM improves the network's focus and overall classification effectiveness. The architecture of CBAM and how we integrate it within the ResNet-50 are shown in Fig.~\ref{fig:pp}.

\subsection{k-NN}
Another distinctive feature of our ResAttention-KNN model is the use of the k-Nearest Neighbors (k-NN)~\cite{guo2003knn} algorithm as the final classifier, which differs from traditional CNN models that typically rely on fully connected (FC) layers. The advantage of this design is that it allows us to perform classification directly in the learned feature space, making the decision process more interpretable and less prone to overfitting—especially beneficial when working with limited datasets. Additionally, k-NN enables us to assess the quality of feature representations without introducing additional trainable parameters.

We also compress the extracted features into a 128-dimensional embedding to complement this classification strategy. This dimensionality reduction helps strike a balance between compactness and representational capacity. Using the original 2048-dimensional outputs from the CNN backbone can introduce redundancy and noise, increasing the risk of overfitting on small datasets. By contrast, projecting the features to a lower-dimensional space forces the network to retain only the most informative components, thereby improving generalization. Moreover, compact embeddings are more compatible with distance-based classifiers like k-NN, which benefit from well-separated, lower-dimensional feature spaces~\cite{chen2020simple, khosla2020supervised}. Embedding sizes in the range of 128–256 have consistently shown strong performance across a variety of metric learning tasks~\cite{musgrave2020metric, schroff2015facenet}, making 128 a practical and well-supported choice in our setting.

To ensure reasonable performance, we systematically examine key hyperparameters of k-NN, including:
\begin{itemize}
\item The number of neighbors ($k$), which determines the classification boundary. In our study, we set $k=5$ based on empirical validation.
\item The distance metric, which influences how similarity between feature vectors is computed. We use the Euclidean distance as it is a standard choice in embedding-based classification.
\end{itemize}
This setup offers a lightweight, interpretable, and effective framework for evaluating the quality of the features learned by our attention-augmented ResNet-50 model.

\section{Experiments}
\label{sec:exp}
In this section, we elaborate on the training and testing procedures, hyperparameter selection, and the experimental results of our proposed models.

\subsection{Training and Testing Procedures}
We initialize the ResNet-50 backbone with pretrained ImageNet weights to leverage general visual knowledge. The entire network, including the CBAM attention modules, is then fine-tuned on the Taxol image dataset using a standard cross-entropy loss. Fine-tuning allows the model to adapt the generic feature representations to the domain-specific morphological patterns found in our dataset.

After this stage, the backbone is frozen and used as a fixed feature extractor. Each input image is transformed into a 128-dimensional embedding via the final projection layer. These embeddings are then classified using a k-NN classifier, which operates in the learned feature space. This decoupled approach separates representation learning from classification, providing a more interpretable and modular framework for evaluating the learned features.

This training and testing pipeline combines the strengths of transfer learning and instance-based classification to improve performance and generalizability on small biomedical datasets.

\subsection{Hyperparameters}

The hyperparameters used for training and evaluation are summarized in Table~\ref{tab:hyperparams}. These values were selected based on prior literature and empirical tuning to balance learning efficiency and generalization.

\begin{table}[htbp]
\centering
\caption{Summary of key hyperparameters used in model training.}
\label{tab:hyperparams}
\renewcommand{\arraystretch}{1.2}
\begin{tabular}{|l|l|}
\hline
\textbf{Hyperparameter} & \textbf{Value} \\\hline
Loss function         & Cross-entropy \\\hline
Optimizer             & SGD with momentum \\\hline
Momentum              & 0.9 \\\hline
Weight decay          & $5 \times 10^{-4}$ \\\hline
Learning rate         & 0.001 \\\hline
Batch size            & 8 \\\hline
Epochs (max)          & 200 \\\hline
Early stopping        & 20 epochs \\\hline
Embedding size        & 128 \\\hline
k in k-NN             & 5 \\\hline
Distance metric       & Euclidean \\
\hline
\end{tabular}
\end{table}

\subsection{Experimental Results}
\begin{table*}[htbp]
\centering
\caption{Macro-averaged precision, recall, F1 score, and accuracy for all models and evaluation strategies. For each metric, we \textbf{bold} the best and \underline{underline} the second-best result.}
\label{tab:all_metrics_summary}
\renewcommand{\arraystretch}{1.2}
\begin{tabular}{|l|l|c|c|c|c|}
\hline
\textbf{Model} & \textbf{Eval Strat} & \textbf{Precision} & \textbf{Recall} & \textbf{F1 Score} & \textbf{Acc} \\
\hline
ResAttention-KNN     & k-NN     & \textbf{0.7566} & \textbf{0.7500} & \textbf{0.7512} & \textbf{0.7500} \\ \hline
ResNet+CBAM           & FC Layer & \underline{0.6481} & \underline{0.6406} & \underline{0.6288} & \underline{0.6406} \\\hline
ResNet                & FC Layer & 0.6308 & 0.6250 & 0.6149 & 0.6250 \\\hline
ResNet+KNN            & k-NN     & 0.5810 & 0.5781 & 0.5700 & 0.5781 \\
\hline
\end{tabular}
\end{table*}

\begin{figure*}[htbp]
    \centering
    \includegraphics[width=0.9\linewidth]{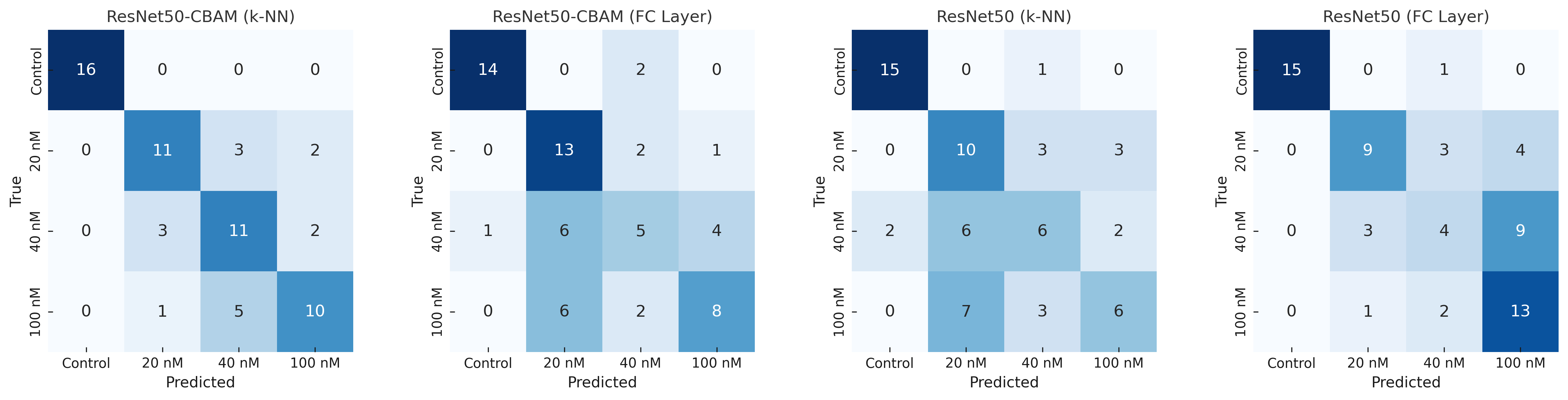}  
    \caption{Confusion matrices for all evaluated models on the Taxol concentration classification task. Each matrix shows the distribution of predicted versus true class labels across four treatment conditions: Control, 20 nM, 40 nM, and 100 nM. }
    \label{fig:cm}
\end{figure*}

We compare our proposed ResAttention-KNN model with several baseline architectures to assess its effectiveness for Taxol concentration classification. Specifically, we evaluate standard ResNet-50 with a fully connected (FC) classification head, ResNet-50 augmented with CBAM, and ResNet-50 using a k-NN classifier on the extracted embeddings. These models represent foundational variants for deep feature learning and instance-based classification. The detailed results are summarized in Table~\ref{tab:all_metrics_summary}.

From the results, it is evident that ResAttention-KNN consistently outperforms all other baselines across all evaluation metrics—precision, recall, F1 score, and accuracy—demonstrating the strength of our model design in morphological classification. Specifically, ResAttention-KNN achieves the highest accuracy of 75.00\%, outperforming the second-best result of 64.06\% obtained by the ResNet+CBAM model with a fully connected layer. It also achieves a macro-averaged F1 score of 0.7512, compared to 0.6288 for the second-best method.

This performance gain highlights the effectiveness of both design choices. The inclusion of CBAM modules leads to improved feature extraction, as observed by the performance gap between ResNet+CBAM (FC) and standard ResNet (FC), which improves from 0.6149 to 0.6288 in F1 score. Meanwhile, swapping the fully connected head with a k-NN classifier further improves the ability to generalize, especially when combined with attention, as seen in the jump from 0.5700 (ResNet+KNN) to 0.7512 (ResAttention-KNN) in F1 score.

These results collectively confirm that each architectural modification—attention integration and k-NN classification—contributes positively to the final performance. Their synergy in the ResAttention-KNN model leads to a significant improvement in classification of Taxol-treated cell images.

\vspace{1em}
\noindent
\textbf{Confusion Matrices. }To further demonstrate that our model effectively captures the morphological distinctions induced by varying Taxol concentrations, we provide the confusion matrices in Fig.~\ref{fig:cm}. From this figure, it is evident that the ResNet50-CBAM (k-NN) model achieves the most accurate predictions, correctly classifying all 16 Control samples and showing improved discrimination among treated groups. For example, among the 20 nM samples, 11 out of 16 were classified correctly, with only minor confusion (3 samples misclassified as 40 nM, and 2 as 100 nM). Similarly, for the 100 nM group, 10 samples were correctly identified, with just a few misclassifications (5 as 40 nM and 1 as 20 nM).

These patterns suggest that the combination of attention-enhanced feature extraction and non-parametric k-NN classification in our ResAttention-KNN model substantially improves class separability, especially in challenging intermediate concentration ranges. This supports the effectiveness of our proposed design for fine-grained morphological classification in limited-data biomedical imaging scenarios.

\section{Conclusion}
\label{sec:con}
In this work, we construct and release a new annotated microscopy image dataset capturing morphological responses of C6 glioma cells to varying concentrations of Taxol. This dataset fills a gap in publicly available resources for automated analysis of chemotherapeutic effects and provides a foundation for future research in biomedical image classification.

Building on this dataset, we design the ResAttention-KNN model, which combines a ResNet-50 backbone with attention modules and a k-NN classifier to improve both representational quality and classification robustness. Experimental results demonstrate that our approach achieves superior performance compared to baseline models, with the highest accuracy (75.00\%) and F1 score (0.7512). The attention mechanism enhances feature discrimination, while the use of k-NN promotes interpretability and effectiveness in data-limited settings.

To support reproducibility and community engagement, we publicly release both the dataset and the complete implementation of our model. We hope these resources will facilitate further research in vision-based drug response analysis and promote the development of more effective biomedical imaging solutions.


\bibliographystyle{IEEEtran}

\end{document}